\PassOptionsToPackage{table}{xcolor}
\documentclass{article}

\usepackage{iclr2026_conference,times}
\iclrfinalcopy
\usepackage{xpeng-defs}
\usepackage{xpeng}

\usepackage{amsfonts}
\usepackage{booktabs}
\usepackage{tabularx}
\usepackage{graphicx}
\usepackage{subcaption}
\usepackage{placeins}
\usepackage{amsmath}
\usepackage{url}
\usepackage{hyperref}
\usepackage{pdfpages}
\hypersetup{
  colorlinks=true,
  linkcolor=xpengblue,
  citecolor=xpengblue,
  urlcolor=xpengblue,
  filecolor=xpengblue,
  pdfborder={0 0 0},
  pdftitle={AnyWorld: Factorized Egocentric World Models for Cross-Embodiment Generalization},
  pdfauthor={Cheng Chen, Jerry Bai, Jiacheng Wei, Boyu Chen, Xiaoji Zheng, Fan Wu, Minghao Yang, Tianrun Chen, Ruibo Li, Xiaoyu Yue, Xiaoyang Guo, Yixiao Ge, Guosheng Lin, Fayao Liu},
}

\setxpenglogo{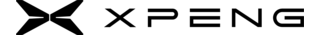}
\setxpenglab{XPENG Robotics}
\setxpengreport{XPENG Technical Report}

\providecommand{\keywords}[1]{\par\noindent\textbf{Keywords:} #1\par}

\title{AnyWorld: Factorized Egocentric World Models \\ for Cross-Embodiment Generalization}

\author{
  \begin{minipage}{0.95\textwidth}
  \centering
  \normalsize\bfseries
  Cheng Chen$^{1,2,3}$, Jerry Bai$^{3,\dagger}$, Jiacheng Wei$^{3}$, Boyu Chen$^{3}$,\\
  Xiaoji Zheng$^{3}$, Fan Wu$^{3}$, Minghao Yang$^{3}$, Tianrun Chen$^{4}$,\\
  Ruibo Li$^{1}$, Xiaoyu Yue$^{3}$, Xiaoyang Guo$^{5}$, Yixiao Ge$^{3}$,\\
  Guosheng Lin$^{1}$, Fayao Liu$^{2,\ddagger}$\\[0.45em]
  {\small\normalfont
  $^{1}$Nanyang Technological University \quad
  $^{2}$Institute of Advanced Intelligence and Computing, A*STAR\\
  $^{3}$XPENG Robotics \quad
  $^{4}$Zhejiang University \quad
  $^{5}$The Chinese University of Hong Kong\\[0.25em]
  {\footnotesize $^{\dagger}$Project Leader \quad $^{\ddagger}$Corresponding Author}\\[0.30em]
  \textbf{Project Page:} \href{https://xpeng-robotics.github.io/anyworld/}{\texttt{xpeng-robotics.github.io/anyworld/}}}
  \end{minipage}
}

\begin{document}

\maketitle

\vspace{-0.25em}
\begin{center}
    \includegraphics[width=0.9\textwidth]{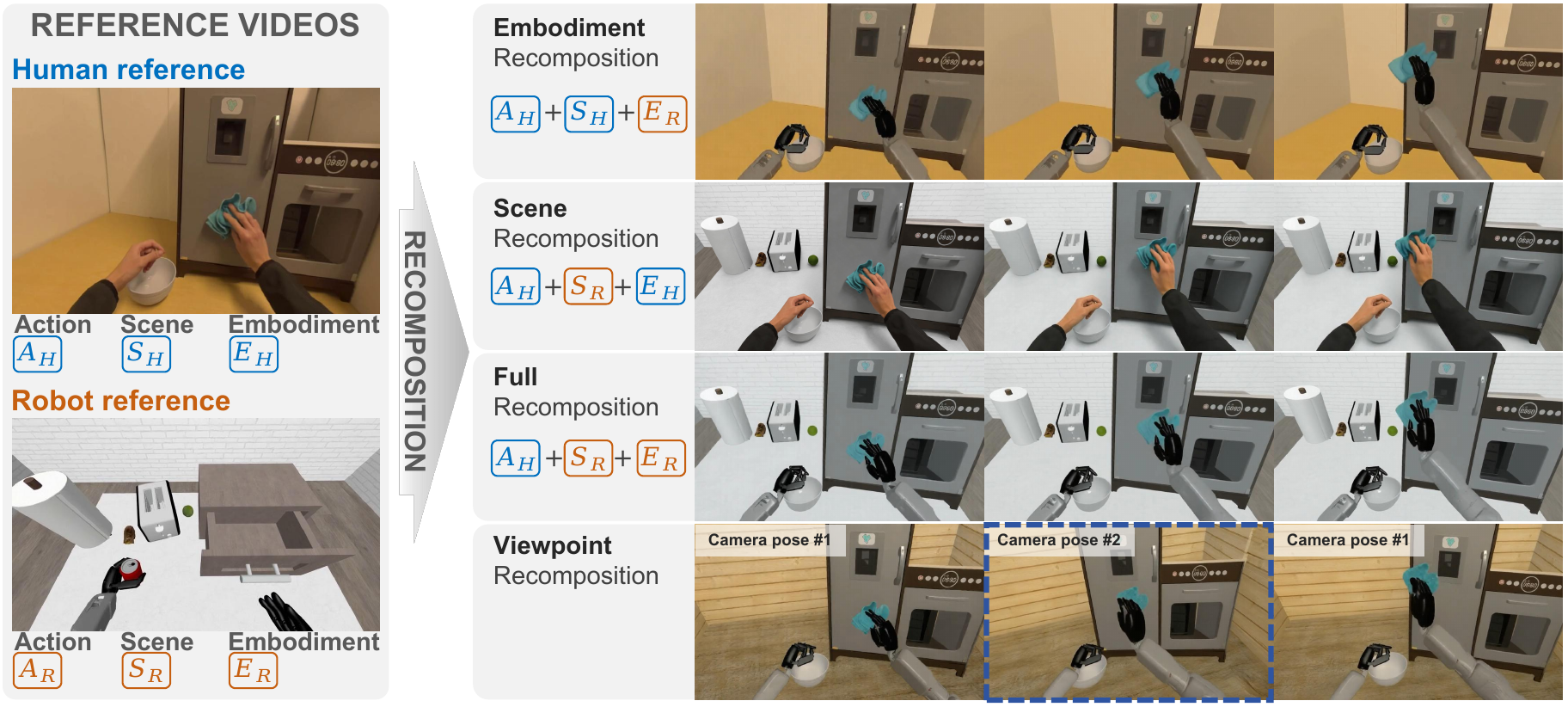}\par
    \includegraphics[width=0.9\textwidth]{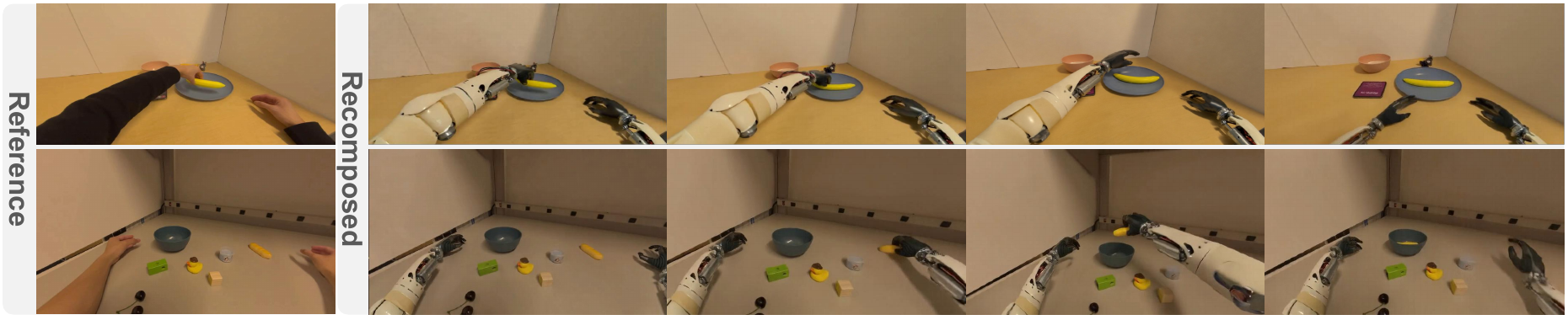}\par
    \captionof{figure}{
    From Human Experience to Diverse Robot Experiences.
    \textbf{Top:} Given a human egocentric interaction, our model generates diverse robot-native rollouts cross embodiments, viewpoints, and scene configurations while preserving the underlying interaction dynamics and object interactions. $A$, $S$, and $E$ denote Action, Scene, and Embodiment, respectively, while subscripts $_H$ and $_R$ indicate human and robot references.
    \textbf{Bottom:} Fine-tuning VLA models on generated real IRON-domain rollouts significantly improves real-robot manipulation robustness.
    }
    \label{fig:teaser}
    \vspace{-0.2em}
\end{center}
\vspace{-0.3em}

\begin{abstract}
Collecting contact-rich robot experiences at scale remains a major bottleneck for generalizable manipulation. Beyond data quantity, robot learning also requires diverse experiences across embodiments, viewpoints, and scenes. Human egocentric videos provide abundant physical interactions, but each video captures only a narrow slice of experience under a single body, camera trajectory, and environment. We propose AnyWorld, a cross-embodiment world modeling framework that expands a single human interaction into diverse robot-native rollouts without paired human-robot demonstrations. Our model factorizes an interaction into action, camera, and embodiment: action controls capture the motion structure, camera controls specify viewpoint evolution, and the target embodiment context defines the acting body and its interaction geometry. This formulation enables independent recomposition of embodiment, viewpoint, and scene factors, allowing a single model to generate many robot-domain experiences while preserving the underlying dynamics and object interactions. We train the model with large-scale human interaction pretraining followed by mixed-embodiment fine-tuning. Experiments show that our model supports controllable recomposition across embodiments, viewpoints, and scenes, and we further demonstrate that the generated data can improve manipulation performance on the RoboCasa GR1 tabletop benchmark and a real IRON humanoid robot. Beyond aggregate gains, we test whether unpaired human experience can be recomposed into robot-native video--action pairs that target a policy gap. Controlled IRON interventions correct a spurious completion prior and establish language-grounded spatial target selection; an action-only counterfactual intervention fails to learn the latter reliably, showing that both action calibration and visual recomposition are necessary.
\end{abstract}

\keywords{Cross-Embodiment World Models, Robot Learning}

\vspace{-0.3em}
\section{Introduction}
\vspace{-0.8em}
Collecting contact-rich robot experience at scale remains a major bottleneck for generalizable manipulation. 
Modern robot learning systems have increasingly benefited from large-scale visual-action datasets~\citep{rt12022,rt22023,openx2024,droid2024,octo2024,openvla2024,pizero2025}, but collecting such data on physical robots remains expensive, hardware-dependent, and difficult to scale across diverse scenes and embodiments. 
Such diversity is not merely a matter of visual augmentation: generalizable policies require not only many interactions, but also repeated exposure to the same task-relevant interaction structures under different bodies, viewpoints, object layouts, and scene contexts.

Human egocentric videos offer a promising source of such experience~\citep{ego4d2022,egoexo4d2024,egodex2025}. 
They are abundant, naturally capture contact-rich manipulation, and cover a wide range of daily physical interactions, which provides useful action variety and auxiliary supervision for robot learning. 
However, human video is still a kind of heterogeneous embodiment data with different visual appearance. 
Furthermore, each individual recording remains tied to a particular configuration. 
Hence, although the underlying interaction may be useful for robot learning, the video itself represents only a narrow slice of experience. 
The central challenge is therefore to treat each recorded interaction as a reusable seed that can be recomposed into multiple robot-native experiences.

This challenge calls for a model that can separate what happens in an interaction from how it is visually realized. 
Here, ``what happens'' refers to the task-relevant interaction, such as motion pattern, object involvement, and temporal progression, while ``how it is realized'' depends on the acting body, viewpoint, and visual scene context. If the motion structure, viewpoint evolution, and acting body can be controlled independently, then a single human interaction can be replayed as multiple robot-domain experiences.
Existing works~\citep{dexmv2021,robotictelekinesis2022,whirl2022,xskill2023,unit2026} have exploited human-robotic transfer, but they do not provide a unified interface that explicitly controls action, camera, and embodiment for producing diverse rollouts. Such a capability would turn human videos from fixed demonstrations into reusable sources of robot training data, allowing the same interaction to be recomposed under different embodiments, cameras, and scene.
We therefore ask: \textit{can a world model expand a single human interaction into diverse robot-native rollouts without paired human-robot demonstrations}?

We propose \textbf{AnyWorld}, an experience recomposition world modeling framework for zero-shot robot experience generation. 
Our key idea is to factorize an interaction into \textit{action, camera, and embodiment conditions}. 
This factorization allows the model to preserve the motion structure and viewpoint evolution of a human interaction while recomposing it under a target robot body and scene context. 
Rather than assuming a fully disentangled scene factor, AnyWorld uses the initial visual state as a target context that specifies the acting body, scene layout, object configuration, and interaction geometry.
As illustrated in Fig.~\ref{fig:teaser}, a single human video can be converted into robot-native rollouts that retain the underlying physical interaction but differ in embodiment, viewpoint, or environment.

AnyWorld is trained by large-scale human interaction pretraining followed by mixed-embodiment fine-tuning.
At inference time, it takes an initial visual state with action, camera, and embodiment conditions to generate target-domain rollouts.
AnyWorld supports body recomposition, viewpoint recomposition, first-frame-based scene recomposition, and their combinations.
These rollouts serve as a scalable synthetic experience multiplier for downstream VLA training.

We evaluate AnyWorld from both world-modeling and robot-learning perspectives. 
First, we show that the model supports controllable recomposition across multiple embodiments, viewpoints, and scenes. 
We quantify whether generated rollouts follow the specified action, camera, and embodiment conditions, and evaluate video quality with standard video-generation metrics.
Second, we use the generated robot-domain rollouts to improve downstream VLA adaptation. 
Experiments in the RoboCasa GR1 tabletop benchmark~\citep{robocasa2024} and on an IRON humanoid robot show that the synthesized experience improves manipulation performance, demonstrating that recomposed robot-domain rollouts can serve as useful training data for target-embodiment adaptation.

In summary, we make three contributions:
(i) we formulate cross-embodiment experience recomposition as a world-modeling problem, enabling a single human egocentric interaction to be expanded into multiple robot-native rollouts without paired human-robot demonstrations;
(ii) we introduce AnyWorld, a factorized video diffusion model that combines action control, camera control, and target embodiment-context conditioning for controllable rollout generation;
and (iii) we demonstrate that the generated robot-domain rollouts improve downstream VLA adaptation in both RoboCasa GR1 simulation and real-robot IRON experiments, and use controlled capability interventions to correct a spurious completion prior and transfer language-grounded spatial target selection. An action-only ablation further establishes the necessity of robot-native visual recomposition.

\vspace{-0.8em}
\section{Related Works}
\vspace{-1em}
\paragraph{Robotic world models.}
World models can turn limited robot data into additional experience by predicting how visual or latent states evolve under actions. Recent robotic world models study action-conditioned video or latent prediction with autoregressive, diffusion, or latent-action interfaces~\citep{ivideogpt,uwm,irasim,adaworld}, and use imagined rollouts for policy learning or improvement~\citep{dreamgen,vlaw,vla_mbpo,dial}. DreamDojo is closest in motivation to ours, as it pretrains a generalist robot world model from large-scale egocentric human videos and transfers interaction priors across embodiments~\citep{dreamdojo}. More broadly, unified multimodal and generative models have explored cross-task representation sharing~\citep{utc2022}, geometry-aware 3D and structured CAD generation~\citep{sculpt3d2024,cadcrafter2025}, and flexible many-to-many image generation using video priors~\citep{imontage2026}. However, prior robot world models typically predict futures within a fixed embodiment domain, adapt to a target robot, or improve action prediction. AnyWorld instead treats one unpaired human interaction as a reusable seed that can be recomposed into many robot-domain experiences by changing action, camera, embodiment, and initial visual context.
\vspace{-0.8em}
\paragraph{Human egocentric videos and cross-embodiment transfer.}
Human egocentric videos provide scalable contact-rich manipulation experience beyond robot teleoperation. Prior work uses them for visual representations and rewards~\citep{r3m2023,vip2023}, reduces visual appearance gaps with masking~\citep{egomimic}, learns latent actions or physical tokens~\citep{lapa,clap,conla,unit2026}, and scales to VLA pretraining or dexterous cross-embodiment transfer~\citep{vitra2025,egodex2025,dexwm,xskill2023,unidex}. Complementary VLA research also improves deployment efficiency and spatiotemporal reasoning in compact policies~\citep{swiftvla2026}. These methods make human experience useful for representation learning, supervision, retargeting, or policy training, but they do not render full robot-native videos from an unpaired human interaction. AnyWorld instead factorizes interaction trajectory, viewpoint evolution, embodiment, and initial visual context, enabling the same human experience to be recomposed across target bodies, cameras, and scene configurations.

\section{Method}
\vspace{-0.4em}
\begin{figure}[t]
\centering
\includegraphics[width=0.95\linewidth]{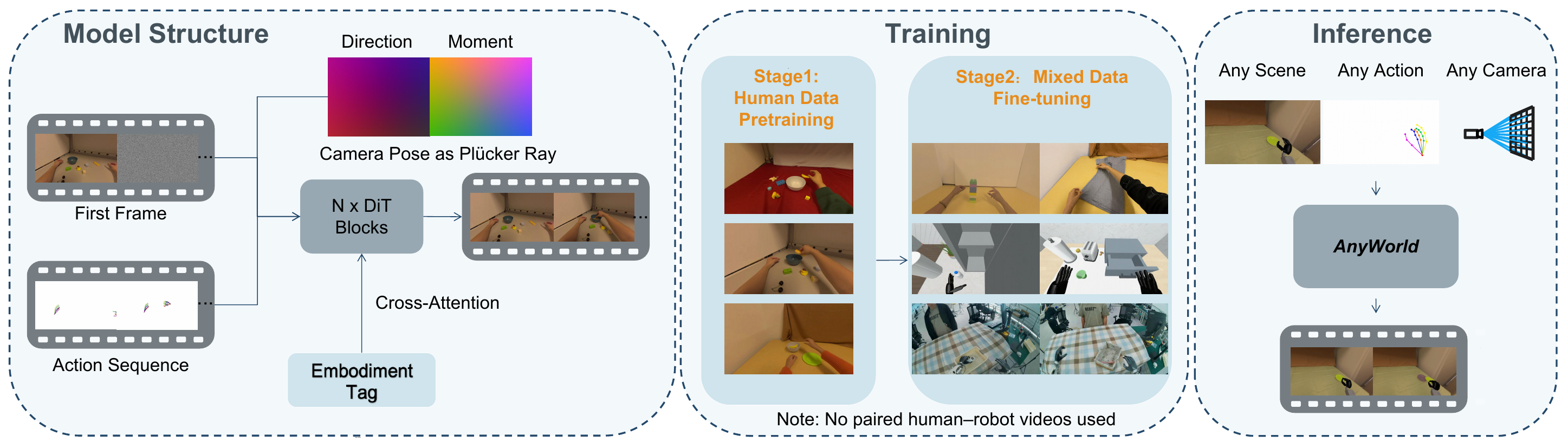}
\caption{
\textbf{Method overview.}
AnyWorld uses action videos, camera controls, and embodiment tags to generate egocentric rollouts.
It is trained by human-video pretraining followed by mixed human--robot fine-tuning, without paired human--robot data.
At inference time, the same model can compose different scenes, actions, cameras, and robot embodiments.
}
\label{fig:method_overview}
\vspace{-1.5em}
\end{figure}

\subsection{Problem Formulation}
\vspace{-0.8em}
We study zero-shot recomposition: converting an egocentric human interaction video into a target-robot visual experience without paired human-robot demonstrations.
Given a source human video \(V^h=\{I^h_0,\ldots,I^h_T\}\), we extract an action control sequence \(A_{1:T}\) and a camera control sequence \(C_{1:T}\).
For a target embodiment \(e\), we obtain an initial frame \(\tilde{I}^e_0\) that specifies the desired body and scene context.
Our goal is to generate
$
    \hat{V}^{e}_{1:T}
    =
    G_{\theta}
    \left(
        \tilde{I}^{e}_0,\,
        A_{1:T},\,
        C_{1:T},\,
        \tau_e
    \right),
$
where \(G_{\theta}\) is a cross-embodiment world model and \(\tau_e\) specifies the target embodiment.
By decoupling action, camera, and embodiment-scene context, the model supports controllable recomposition of interaction trajectory, viewpoint, embodiment, and scene at inference time.
The model never observes paired clips of the same interaction performed by both humans and robots.

\subsection{Action-Camera-Embodiment Factorization}

As shown in Fig.~\ref{fig:method_overview}, we factorize each interaction into three components: action, camera, and embodiment. 
This design exposes the main factors that determine an egocentric manipulation video: what motion is performed, how the scene is observed, and which body performs the interaction.

The action condition \textbf{\(A_{1:T}\)} is represented as a rendered skeleton control video. 
This places the action in the image plane, specifying where motion should occur in the video rather than how a particular body should actuate. Compared with joint commands or robot-specific action vectors, such pixel-space control provides a more embodiment-agnostic interface and is easier for a video model to share across human and robot bodies. The control video is encoded into a latent tensor \(Z^a_{1:T}\), which serves as a dense motion condition for generation. 

The camera condition \(C_{1:T}\) is constructed from camera intrinsics and extrinsics. 
For each frame, we convert the camera parameters into a Pl\"ucker ray embedding. 
For a pixel location \(u\), the ray is represented as
$
    r(u)=\left(d(u),\, o \times d(u)\right),
$
where \(o\) is the camera center and \(d(u)\) is the ray direction in world coordinates. 
By conditioning on camera geometry, the model can better separate actor motion from camera motion. 

The embodiment is specified by the initial frame and a text tag. 
The initial frame provides the target body appearance, scene context, object layout, and spatial relationship at the start of the rollout. 
The text tag \(\tau_e\) identifies the acting embodiment, e.g., human, RoboCasa GR1, or IRON, and is encoded together with the caption. 
Together, the initial frame and text condition allow the same action-camera controls to be recomposed with different bodies and scenes at inference time.

\subsection{Cross-Embodiment World Model}
\vspace{-0.8em}
We implement \(G_{\theta}\) as a latent diffusion video model. 
Given the clean video latent \(Z_0\), noisy latent \(Z_s\), action latent \(Z^a_{1:T}\), camera embedding \(Z^c_{1:T}\), and text embedding \(E_{\tau}\), the model predicts the diffusion noise conditioned on all three factors.

The three conditioning streams are injected in different ways. 
First, the action control video is encoded beforehand and concatenated with the noisy video latent along the channel dimension:
\begin{equation}
    H_{\mathrm{in}}
    =
    \mathrm{Concat}
    \left[
        Z_s,\,
        Z^a_{1:T},\,
        Z^0_{1:T}
    \right],
\end{equation}
where \(Z^0_{1:T}\) is a zero latent used for input-format compatibility. 
Thus, the action control enters the DiT as an input-channel condition.

Second, the camera control is packed into \(Z^c_{1:T}\). 
It is passed through a lightweight adapter and added to the patch embedding:
\begin{equation}
    H_0
    =
    \mathrm{PatchEmbed}
    \left(
        H_{\mathrm{in}}
    \right)
    +
    \mathrm{Adapter}
    \left(
        Z^c_{1:T}
    \right).
\end{equation}
This provides an additive geometric conditioning signal for viewpoint evolution. Third, the embodiment tag and caption are encoded as text embeddings \(E_{\tau}\). 
They are injected into each transformer block through cross-attention:
\begin{equation}
    H_{\ell+1}
    =
    \mathrm{DiTBlock}_{\ell}
    \left(
        H_{\ell};\,
        E_{\tau}
    \right).
\end{equation}
This allows the same model to bind the rollout to different embodiment domains.

\subsection{Training}
\vspace{-0.8em}
Training consists of two stages. 
First, we pretrain the model on large-scale egocentric human interaction videos from EgoDex. 
This stage exposes the model to diverse contact-rich motion patterns and learns a strong action-camera conditioned video prior. Second, we fine-tune the model on unpaired multi-embodiment data. 
Each video, whether from a human or robot embodiment, is converted into the same factorized format: action control, camera control, and embodiment condition. 
No human-robot clip-level correspondence is used. 
Because all embodiments share the same action-camera conditioning interface, the model learns to associate different embodiment appearances with compatible interaction dynamics.

\subsection{Zero-Shot Experience Recomposition}
\vspace{-0.8em}
At inference time, the factorized interface allows us to recompose a new robot experience from independently specified factors.
Given a source video, we extract its action and camera controls \((A_{1:T}, C_{1:T})\).
We then edit only the initial frame to specify the target visual context, obtaining \(\tilde{I}^{e}_0\).
This edited frame can change the acting body, background scene, or visual style while preserving the spatial layout and interaction objects.
Together with the embodiment tag \(\tau_e\), it defines the desired target domain for rollout generation. The generated rollout is obtained by
$
    \hat{V}^{e}_{1:T}
    =
    G_{\theta}
    \left(
        \tilde{I}^{e}_0,\,
        A_{1:T},\,
        C_{1:T},\,
        \tau_e
    \right).
$ This formulation supports several forms of zero-shot recomposition under the same model.
By editing only \(\tilde{I}^{e}_0\) and changing \(\tau_e\), the same human action-camera trajectory can be rendered with a robot body, a new scene, or both.
By changing \(C_{1:T}\) while keeping the action and edited initial frame fixed, the model can generate the same interaction under a different viewpoint.

{
\subsection{Joint Visual--Action Experience Recomposition}
\label{sec:visual_action_pair}
For downstream policy training, we transfer both the observation and the action rather than attaching robot actions to unmodified human imagery. AnyWorld converts the source video into a target-robot observation by re-embodying the actor and, when needed, intervening on the task-relevant scene state. In parallel, an action-calibration module maps the human wrist trajectory into the target robot's relative end-effector action space. Each transferred experience is therefore
\begin{equation}
    \mathcal{D}_{\mathrm{trans}}
    =
    \left\{
    \left(
        \hat{V}^{e}_{1:T},\,
        \ell,\,
        \widetilde{A}^{e}_{1:T}
    \right)
    \right\},
    \label{eq:visual_action_pair}
\end{equation}
where $\hat{V}^{e}_{1:T}$ is the re-embodied and optionally re-scened rollout, $\ell$ is the task instruction, and $\widetilde{A}^{e}_{1:T}$ is the morphology-calibrated robot action. The two transferred components play different roles: action calibration specifies what the robot should execute, whereas visual recomposition supplies the robot-native state in which the instruction and action must be grounded. Section~\ref{sec:targeted_policy} evaluates these components through controlled closed-loop interventions. Detailed action conversion and data construction are provided in the supplementary material.
}

\section{Experiments}
\vspace{-0.8em}
We evaluate AnyWorld from two perspectives: controllable experience recomposition and downstream VLA adaptation.
Specifically, we ask whether AnyWorld can (i) follow the specified recomposition controls; (ii) generate different recomposition types, including body, viewpoint, and scene recomposition; (iii) provide useful robot-domain experience for policy adaptation; and, beyond average gains, (iv) recompose unpaired human experience into robot-native video--action pairs that selectively fill a pre-identified policy capability gap.

\subsection{Experimental Setup}
\paragraph{Data and training.}
We use EgoDex~\citep{egodex2025} as the source of human egocentric interaction data, and use RoboCasa GR1~\citep{robocasa2024} and IRON videos as robot-domain data.
All data is converted into the same factorized interface: an action control video, a camera trajectory represented by Pl\"ucker rays, and an embodiment tag.
Our world model is initialized from WAN Fun-Control 14B~\citep{wanfuncontrol2025}.
We pretrain on 200K EgoDex clips for 30K steps to learn action-camera conditioned human interaction priors, and then perform mixed-embodiment fine-tuning for 5K steps using unpaired EgoDex, RoboCasa GR1, and IRON clips.
For each target robot domain, we use 5K EgoDex clips and 5K robot-domain clips.
No paired human-robot videos are used at any stage.

\paragraph{Baselines.}
We compare against two strong baselines.
WAN Fun-Control is a large-scale control-video conditioned generation model based on WAN 2.1 14B~\citep{wanfuncontrol2025}, which provides strong action-control priors.
Cosmos-Predict2.5 is a 14B world foundation model designed for physical world prediction and video generation~\citep{cosmospredict25}; we use its video-to-world generation setting.

\paragraph{Generation metrics.}
Zero-shot recomposition does not provide paired robot ground-truth videos, so we evaluate factor controllability with automatic proxy metrics.
\textbf{ActionAlign} compares the motion of the generated foreground region with the motion of the input action-control video using optical-flow similarity within the control mask.
\textbf{CameraAlign} compares the background optical flow of the generated video with that of the source video clip, which serves as a proxy for the input camera trajectory.
\textbf{EmbodAcc} is a CLIP-based binary accuracy that checks whether generated frames are closer to target-embodiment prompts than to alternative embodiment prompts.
For video quality, we also report standard VBench~\citep{vbench2024} metrics, including subject consistency, background consistency, flickering, and smoothness.
Metric details are included in the supplementary material.

\subsection{Qualitative Robot Experience Recomposition}

\begin{figure}[t]
    \centering
    \begin{subfigure}{0.98\linewidth}
        \centering
        \includegraphics[width=\linewidth]{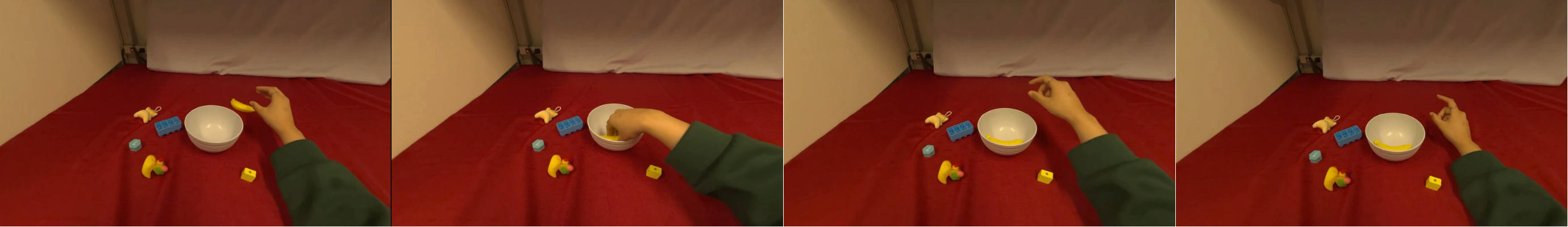}
        \caption{Human (Reference)}
        \label{fig:zeroshot-human}
    \end{subfigure}
    \begin{subfigure}{0.98\linewidth}
        \centering
        \includegraphics[width=\linewidth]{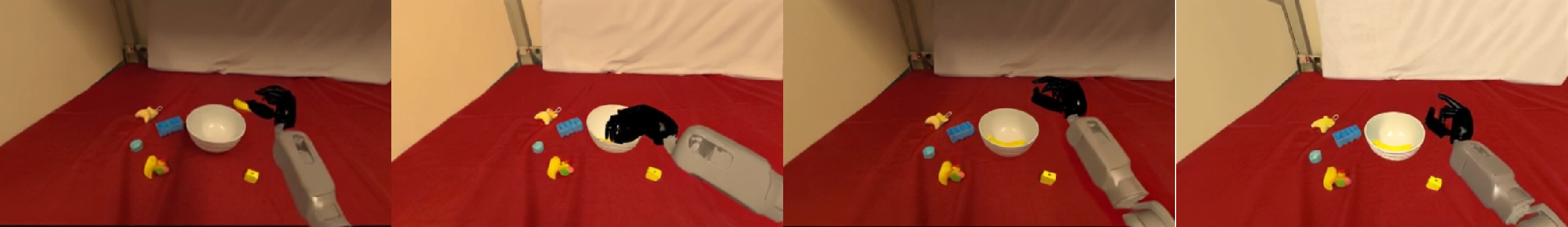}
        \caption{Robocasa GR1 (Recomposed)}
        \label{fig:zeroshot-gr1}
    \end{subfigure}
    \begin{subfigure}{0.98\linewidth}
        \centering
        \includegraphics[width=\linewidth]{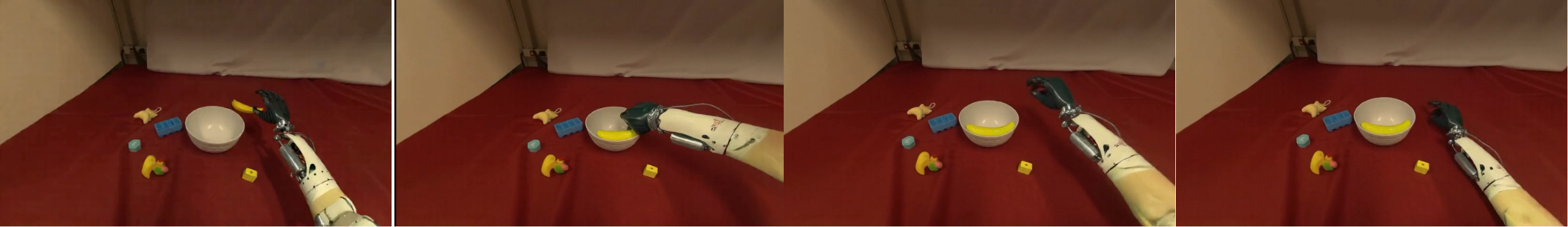}
        \caption{IRON (Recomposed)}
        \label{fig:body_recomposition}
    \end{subfigure}
    \caption{
    \textbf{Zero-Shot Recomposition.} Given the same action and camera controls extracted from a source human interaction, our model changes the acting body to RoboCasa GR1 or IRON while preserving the interaction structure.
    }
    \label{fig:same_interaction}
\end{figure}

\paragraph{Embodiment recomposition.}
We first show that the model can recompose the same human interaction under different robot bodies.
Given an EgoDex human clip, we keep the action and camera controls fixed, and change only the target initial frame and embodiment tag.
As shown in Fig.~\ref{fig:same_interaction}, the model can instantiate the same interaction using RoboCasa GR1 and IRON embodiments while preserving the source motion and scene layout.

\begin{figure}[t]
\vspace{-0.8em}
\centering

\includegraphics[width=0.9\linewidth]{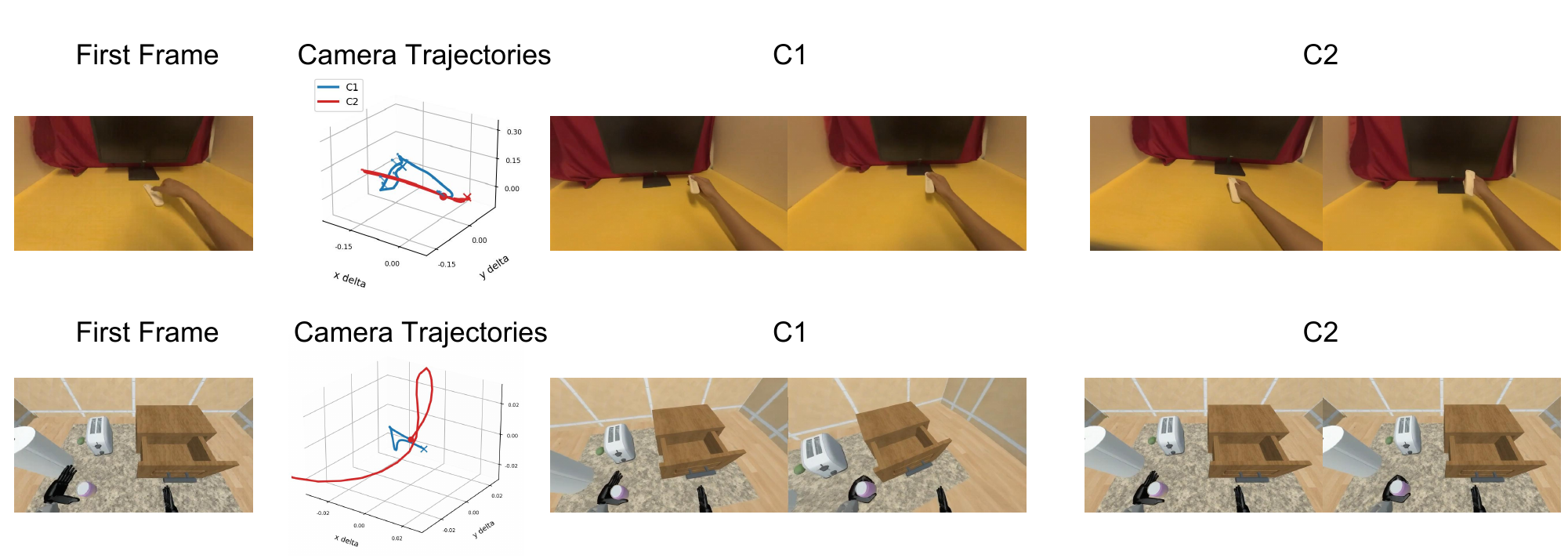}
\vspace{-0.4em}
\caption{
\textbf{Viewpoint recomposition.}
With the action, body, and scene fixed, changing the camera condition leads to a different viewpoint while maintaining the same interaction.
}
\label{fig:viewpoint_recomposition}
\vspace{-0.8em}
\end{figure}

\paragraph{Viewpoint recomposition.}
We next test camera controllability.
We keep the same action, target body, and scene, but change the camera trajectory.
Fig.~\ref{fig:viewpoint_recomposition} shows that the model changes the viewpoint while preserving the same physical interaction, indicating that camera control is not merely absorbed into the action condition. This is important for egocentric robot learning, where apparent object and hand motion are strongly coupled with camera motion.

\begin{figure}[t]
    \centering
    \includegraphics[width=0.85\linewidth]{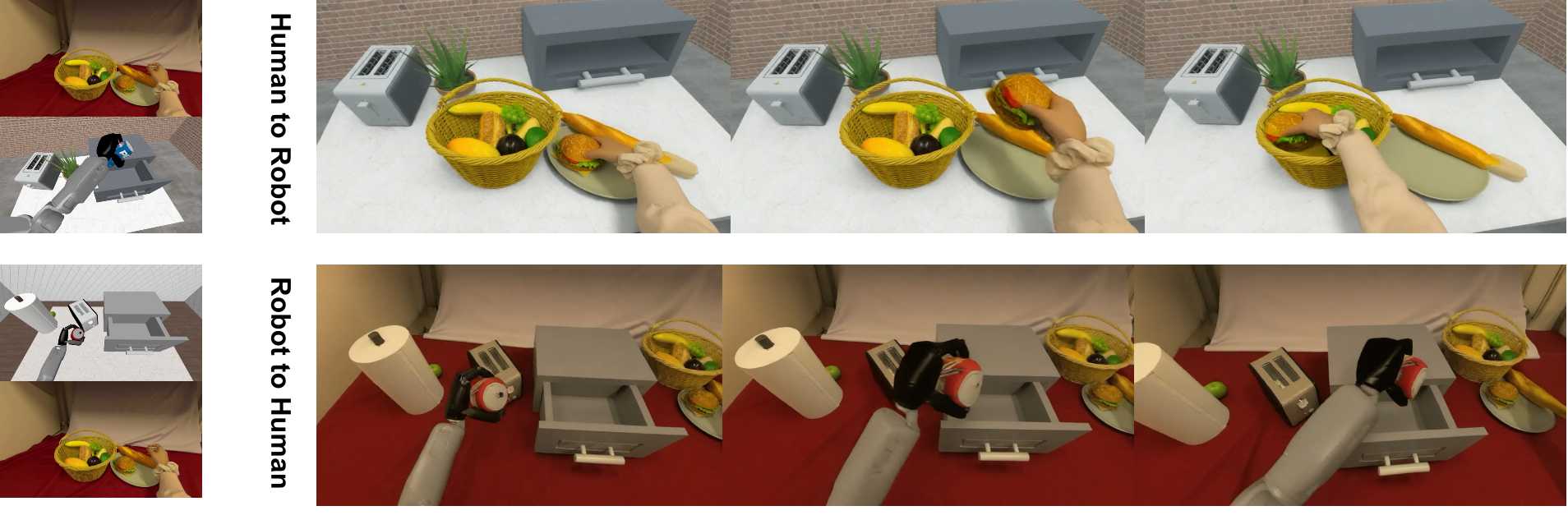}
    \vspace{-0.5em}
    \caption{
    \textbf{Scene and body recomposition.}
    Our model reuses the same action-camera controls while recomposing the acting body and scene.
    The top row shows human-to-RoboCasa GR1 recomposition, and the bottom row shows RoboCasa GR1-to-human recomposition.
    }
    \label{fig:scene_body_recomposition}
    \vspace{-1em}
\end{figure}

\paragraph{Scene and body recomposition.}
Finally, we demonstrate first-frame based scene and body editing.
The model keeps the human action-camera controls while changing the acting body, the background scene, or both.
This allows a single human interaction to be recomposed into different robot-domain visual experiences, as shown in Fig.~\ref{fig:scene_body_recomposition}.
Together, these examples illustrate the core capability of AnyWorld: one recorded interaction can serve as a seed for multiple robot-native rollouts.

\subsection{Controllability and Video Quality}
\vspace{-0.8em}
Generated rollouts must be both visually plausible and controllable.
We evaluate controllability on 60 videos across IRON, RoboCasa GR1, and EgoDex using \textit{ActionAlign}, \textit{CameraAlign}, and \textit{EmbodAcc}, and report their average as an overall score.

As shown in Table~\ref{tab:control_metrics}, Cosmos-Predict2.5 is weakly grounded in the specified controls, while WAN Fun-Control follows actions well but lacks explicit camera modeling.
Our factorized model achieves the best action, camera, and embodiment controllability, demonstrating the benefit of the proposed action-camera-embodiment factorization.

\begin{table}[t]
    \centering
    \caption{
    \textbf{World-model controllability.}
    We evaluate whether generated rollouts follow the specified action, camera, and embodiment conditions.
    }
    \label{tab:control_metrics}

    \begingroup
    \setlength{\tabcolsep}{16pt}
    \renewcommand{\arraystretch}{1.05}

    \resizebox{0.9\linewidth}{!}{
    \begin{tabular}{lcccc}
        \toprule
        Method
        & ActionAlign $\uparrow$
        & CameraAlign $\uparrow$
        & EmbodAcc $\uparrow$
        & Avg. $\uparrow$ \\
        \midrule
        Cosmos-Predict2.5
        & 0.170 & 0.315 & 0.765 & 0.417 \\
        WAN Fun-Control
        & 0.655 & 0.402 & 0.769 & 0.609 \\
        \rowcolor{gray!10}
        \textbf{AnyWorld (Ours)}
        & \textbf{0.659} & \textbf{0.789} & \textbf{0.886} & \textbf{0.778} \\
        \bottomrule
    \end{tabular}
    }

    \endgroup
\end{table}

We further evaluate video quality using standard VBench metrics.
Table~\ref{tab:vbench_quality} shows that our model maintains strong subject consistency, background consistency, flicker stability, and motion smoothness.
Importantly, the gains in Table~\ref{tab:control_metrics} are not obtained at the cost of video quality: our method remains competitive with, and slightly improves over, the baselines on the average VBench score.

\begin{table}[t]
    \centering
    \caption{
    \textbf{Video quality evaluation.}
    We report VBench subject consistency, background consistency, flickering, and smoothness.
    Avg. is computed over the four VBench metrics.
    }
    \label{tab:vbench_quality}
    \setlength{\tabcolsep}{12pt}
    \resizebox{0.9\linewidth}{!}{
    \begin{tabular}{lccccc}
        \toprule
        Method
        & Subject $\uparrow$
        & Background $\uparrow$
        & Flicker $\uparrow$
        & Smoothness $\uparrow$
        & Avg. $\uparrow$ \\
        \midrule
        Cosmos-Predict2.5
        & 0.923 & 0.940 & 0.990 & 0.994 & 0.962 \\
        WAN Fun-Control
        & 0.937 & 0.946 & 0.993 & \textbf{0.996} & 0.968 \\
        \rowcolor{gray!10}
        \textbf{AnyWorld (Ours)}
        & \textbf{0.942} & \textbf{0.949} & \textbf{0.995} & \textbf{0.996} & \textbf{0.971} \\
        \bottomrule
    \end{tabular}
    }
    \vspace{-0.5em}
\end{table}

\begin{table}[h]
\vspace{-0.8em}
    \centering
    \caption{
    \textbf{Ablation on human-to-robot ratio.}
    We evaluate mixed fine-tuning ratios under the full re-embodiment setting.
    Avg. is computed over ActionAlign, CameraAlign, and EmbodAcc.
    }
    \label{tab:ratio_ablation}

    \begingroup
    \setlength{\tabcolsep}{14pt}
    \renewcommand{\arraystretch}{1.05}

    \resizebox{0.8\linewidth}{!}{
    \begin{tabular}{lcccc}
        \toprule
        Human:Robot
        & ActionAlign $\uparrow$
        & CameraAlign $\uparrow$
        & EmbodAcc $\uparrow$
        & Avg. $\uparrow$ \\
        \midrule
        4:1
        & 0.601 & 0.727 & 0.789 & 0.706 \\
        1:1
        & 0.603 & \textbf{0.775} & 0.791 & 0.723 \\
        \rowcolor{gray!10}
        2:1
        & \textbf{0.623} & 0.754 & \textbf{0.842} & \textbf{0.740} \\
        \bottomrule
    \end{tabular}
    }

    \endgroup
    \vspace{0.15em}
\end{table}

\begin{table}[!t]
    \centering
    \caption{
    \textbf{Transferred experience for VLA adaptation.} Compared to the VLA baseline, we add our re-embodied EgoDex rollouts during target-embodiment adaptation.
    }
    \label{tab:vla_adaptation}

    \begingroup
    \setlength{\tabcolsep}{14pt}
    \renewcommand{\arraystretch}{1.05}
    \footnotesize
    \resizebox{0.9\linewidth}{!}{
    \begin{tabular}{lcccc}
        \toprule
        Setting
        & Evaluation
        & Baseline
        & + Ours
        & Gain \\
        \midrule
        RoboCasa GR1
        & 18 Pick-and-Place Tasks
        & 49.8\%
        & \textbf{54.6\%}
        & +4.8 \\
        IRON Real Robot
        & 20 Grasp Trials
        & 20.0\%
        & \textbf{55.0\%}
        & +35.0 \\
        \midrule
        Macro Avg.
        & --
        & 34.9\%
        & \textbf{54.8\%}
        & +19.9 \\
        \bottomrule
    \end{tabular}
    }

    \endgroup
    \vspace{-0.2em}
\end{table}

\subsection{Training Ablations}
\vspace{-0.8em}
We study the effect of the human-to-robot data ratio during mixed fine-tuning under the full recomposition setting.
For a controlled comparison, all ratios are evaluated on the same 60 held-out bidirectional cases: 30 EgoDex-to-RoboCasa GR1 and 30 RoboCasa GR1-to-EgoDex videos.
As shown in Table~\ref{tab:ratio_ablation}, the 2:1 mixture achieves the best average controllability over ActionAlign, CameraAlign, and EmbodAcc.

The 4:1 mixture provides richer human interaction patterns but weaker robot-domain grounding, leading to lower CameraAlign and EmbodAcc.
The 1:1 mixture improves CameraAlign, likely due to stronger exposure to robot-domain viewpoints, but still yields lower EmbodAcc than 2:1.
This suggests that cross-embodiment generation requires balancing human action-camera diversity with robot embodiment grounding, rather than simply increasing either side.
We therefore use the 2:1 mixture for our main model.

\subsection{Transferred Experience for VLA Adaptation}

We evaluate whether generated robot-domain rollouts improve downstream VLA deployment. We construct the VLA baseline following UniT~\citep{unit2026}, which already uses large-scale EgoDex data during pretraining. This makes the setting stringent: we test whether \emph{re-embodied} human experience still provides gains during target-embodiment adaptation. Following the common pretraining-then-adaptation paradigm for VLA systems~\citep{openx2024,unit2026}, we add our generated data only in the second stage. Given EgoDex clips, we preserve their action-camera trajectories and recompose them into the target robot domain: GR1 in RoboCasa simulation and IRON for real-world deployment. During adaptation, transferred EgoDex rollouts are mixed with target-domain robot data at a 1:1 ratio, providing robot-domain visual experience synthesized from human physical interactions. Table~\ref{tab:vla_adaptation} reports the results. On 18 RoboCasa GR1 pick-and-place tasks, our transferred experience improves UniT from 49.8\% to 54.6\%. On the real IRON robot, the same strategy improves success from 20.0\% to 55.0\% over 20 banana-grasping trials with varying backgrounds. These results show that AnyWorld serves as a deployment-stage data engine that converts human physical experience into target-robot observations for policy adaptation.

\begin{figure}[!htbp]
    \centering
    \includegraphics[width=0.96\linewidth]{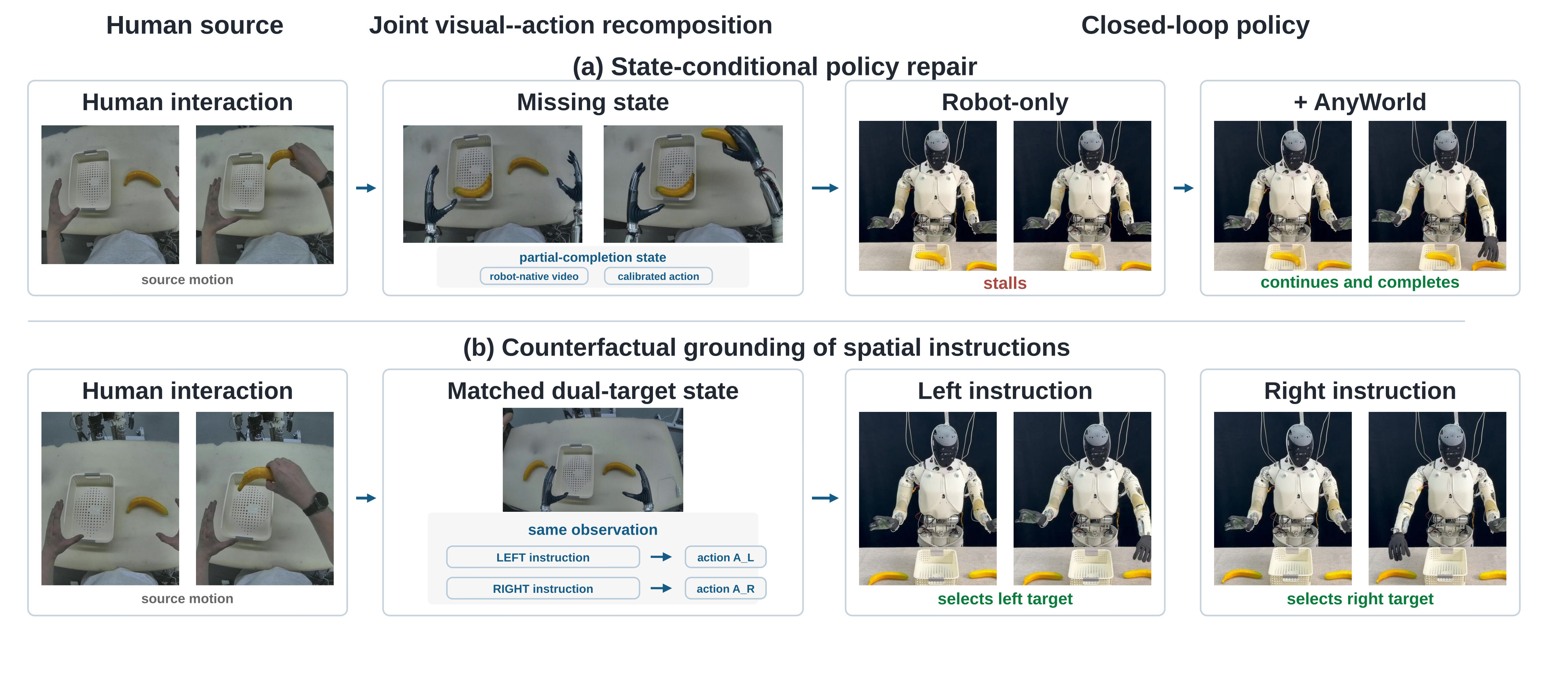}
    \caption{\textbf{Targeted capability transfer from unpaired human experience.} We jointly recompose human interaction in observation and action to fill a known robot-policy gap. \textbf{(a) State-conditional policy repair.} The baseline possesses the underlying pick-and-place skill but stalls in a partial-completion state absent from robot training. Re-scened IRON observations paired with calibrated human actions restore task continuation. \textbf{(b) Counterfactual grounding of spatial instructions.} AnyWorld creates a matched dual-target robot-native observation and pairs it with opposite spatial instructions and corresponding actions. The adapted policy changes the selected target with the instruction.}
    \label{fig:targeted_policy}
    \vspace{-0.6em}
\end{figure}

{
\subsection{Targeted Capability Transfer through Controlled Recomposition}
\label{sec:targeted_policy}
Beyond aggregate adaptation gains, we test a narrower and falsifiable proposition: \emph{without paired human--robot data, can human experience be recomposed into robot-native video--action pairs that target a pre-identified policy capability gap?} Each intervention begins with a closed-loop failure whose missing behavior is known in advance. We then add only the hypothesized missing experience while keeping the policy architecture, robot data, and adaptation protocol unchanged. The two pick-and-place instantiations below separately test state coverage and language-grounded control.

\paragraph{State-conditional policy repair.}
The robot-only policy already succeeds in the standard single-target setting, establishing that its basic reach--grasp--place capability is available. However, in a partially completed state---one target is already inside the receptacle while another remains outside---the policy becomes nearly stationary. This failure reveals a spurious completion prior rather than a motor deficit. We therefore re-scene human demonstrations into the missing robot-domain state and pair them with calibrated IRON actions. After adaptation, the same policy resumes task-directed motion and completes the remaining placement, as shown in Fig.~\ref{fig:targeted_policy}(a). The controlled change isolates missing state coverage as the relevant intervention.

\paragraph{Counterfactual grounding of spatial instructions.}
The original robot data contains single-target, single-side trajectories and therefore never requires language to disambiguate symmetric alternatives. AnyWorld constructs a matched dual-target IRON observation and pairs the same visual state with opposite spatial instructions and their corresponding left/right actions. The scene and task are held fixed; only the linguistic target and aligned action change. Crucially, an action-only intervention---counterfactual left/right actions paired with the original one-sided IRON observations---does not produce reliable instruction following. Stable target switching appears only when the counterfactual actions are grounded in the re-embodied and re-scened human visual pairs. This ablation shows that calibrated motion is useful but not sufficient: visual recomposition supplies the missing robot-native state on which language and action can be jointly grounded.
}

\begin{table}[!htbp]
    \centering
    \caption{\textbf{Controlled visual--action intervention audit.} The action-only setting adds counterfactual left/right actions while retaining the original one-sided IRON observations. Only joint visual--action recomposition supplies both the missing robot-native state and aligned action supervision.}
    \label{tab:targeted_intervention_audit}
    \begingroup
    \small
    \setlength{\tabcolsep}{5pt}
    \renewcommand{\arraystretch}{1.12}
    \begin{tabularx}{\linewidth}{l>{\raggedright\arraybackslash}X>{\raggedright\arraybackslash}X>{\raggedright\arraybackslash}X}
        \toprule
        Setting
        & Robot-native visual coverage
        & Action coverage
        & Closed-loop evidence \\
        \midrule
        Robot-only baseline
        & Standard single-target states only
        & Original robot actions
        & Stalls in the partial-completion state; no spatial instruction grounding \\
        + Counterfactual actions only
        & Unchanged one-sided robot observations
        & Paired left/right actions
        & Spatial instruction following remains unstable \\
        \rowcolor{gray!10}
        \textbf{+ AnyWorld joint recomposition}
        & Partial-completion and matched dual-target states
        & Calibrated human actions and paired left/right actions
        & \textbf{Completes the remaining placement and stably follows spatial instructions} \\
        \bottomrule
    \end{tabularx}
    \endgroup
    \vspace{-0.6em}
\end{table}

{
Together, these experiments provide controlled evidence for a data-expansion mechanism rather than a generic scaling effect. The baseline deficiency is identified before intervention, the added visual and action factors are explicit, and the resulting policy change is evaluated in closed loop. AnyWorld therefore turns unpaired human experience into targeted robot-native supervision that can either repair a known state-dependent failure or transfer a language-grounded capability absent from the robot data.
}
\FloatBarrier
\vspace{-0.8em}

\section{Conclusion}
\vspace{-0.8em}
We presented \textbf{AnyWorld}, a factorized egocentric world model for cross-embodiment manipulation.
The key idea is to decompose an interaction into action, camera, and embodiment, enabling a single model to recompose human experience into robot-domain rollouts without paired human-robot demonstrations.
By combining large-scale EgoDex pretraining with unpaired mixed-embodiment fine-tuning, our model supports controllable recomposition across bodies, viewpoints, and scenes while preserving the underlying interaction structure.
Experiments show that AnyWorld improves action-camera-embodiment controllability over strong video/world-model baselines and maintains competitive video quality.
More importantly, the generated robot-domain experience improves downstream UniT-based VLA adaptation in RoboCasa GR1 simulation and on a real IRON robot. Controlled capability interventions further repair a spurious completion prior and transfer language-grounded spatial target selection. The failure of action-only counterfactual pairing shows that this effect is not explained by action relabeling alone: robot-native visual recomposition is required to ground the transferred behavior.
These results suggest that cross-embodiment world models can serve as scalable experience engines for robot learning, turning abundant human interaction videos into useful robot-native training data.

\vspace{-0.8em}
\section{Limitations}
\vspace{-0.8em}
AnyWorld recomposes human egocentric interactions into robot-domain visual experience, but visual rollouts cannot fully capture tactile feedback, contact forces required for precise contact-rich control. The method also depends on reliable action-camera extraction; severe occlusion, fast motion, tracking errors, or strong camera shake may reduce controllability.
Finally, our current evaluation covers 3 embodiments, leaving broader generalization to more robot morphologies, objects, and long-horizon tasks for future work.
Future directions include tactile supervision, uncertainty-aware filtering, and larger mixed-embodiment datasets.

\begingroup
\small
\setlength{\bibsep}{1.5pt}
\bibliography{references}
\bibliographystyle{iclr2026_conference}
\endgroup

\clearpage


\section*{Supplementary material}

# Supplementary Material for AnyWorld: Factorized Egocentric World Models for Cross-Embodiment Generalization

**Cheng Chen**[1,2,3], **Jerry Bai**[3,†], **Jiacheng Wei**[3], **Boyu Chen**[3], **Xiaoji Zheng**[3], **Fan Wu**[3], **Minghao Yang**[3], **Tianrun Chen**[4], **Ruibo Li**[1], **Xiaoyu Yue**[3], **Xiaoyang Guo**[5], **Yixiao Ge**[3], **Guosheng Lin**[1], **Fayao Liu**[2,‡]

[1]Nanyang Technological University [2]Institute of Advanced Intelligence and Computing, A*STAR
[3]XPENG Robotics [4]Zhejiang University [5]The Chinese University of Hong Kong

[†]Project Leader [‡]Corresponding Author

## 1 Overview of Supplementary Materials

This supplementary material provides additional implementation details, qualitative results, metric definitions, and downstream adaptation analyses for AnyWorld. In particular, we describe the reverse pseudo-pair construction used for human-to-robot embodiment editing, the mixed-embodiment training recipe and structured conditioning, policy-adaptation details, quantitative metric definitions, and additional evaluations of interaction outcomes under motion variations.

## 2 Implementation Details

### 2.1 Generalizable Human-to-Robot Embodiment Editing via Reverse Pseudo-Pairs

The first-frame visual condition is critical for cross-embodiment rollout generation, as it specifies the target body appearance, object configuration, scene layout, and initial interaction geometry. To instantiate a target robot embodiment in the first frame of a human egocentric video, we train an embodiment image editor that replaces visible human arms, hands, and body parts with a specified robot morphology while preserving the background, manipulated objects, and camera geometry. We use the same pipeline for multiple target embodiments, including IRON and RoboCasa GR1 (Nasiriany et al., 2024). The procedure is not tied to a specific robot body and can be extended to other embodiments as long as target-embodiment images are available.

A key challenge is that paired human–robot images with identical scene layout, object state, contact configuration, and camera viewpoint are not available. Directly collecting such pairs would require reproducing the same interaction state across different embodiments, which is impractical and would undermine the zero-shot nature of our experience recomposition setting. We therefore introduce a *reverse pseudo-pair construction* strategy. The key observation is that open-domain image-editing models (Google, 2025) are usually much stronger at generating plausible human hands and arms than at generating embodiment-specific robot morphology. We exploit this asymmetry by using the direction in which the foundation editor is strongest: instead of directly editing human images into robot images, we first edit real robot images into human-like observations. The original robot image and the generated human-like image then form an approximately aligned pseudo pair, which provides supervision for training the inverse human-to-robot editor required by AnyWorld.

#### 2.1.1 Reverse pseudo-pair construction

For each target embodiment $e$, such as IRON or RoboCasa GR1, we sample approximately 800 egocentric robot frames from the corresponding robot-domain dataset. Each frame contains the target robot embodiment, the manipulated object, and the surrounding scene. We then use an off-the-shelf large-scale image-editing model, Nano-Banana (Google, 2025) in our implementation, to perform robot-to-human editing. The editing instruction asks the model to preserve the camera pose, background, object config-

uration, and interaction state, while replacing only the visible robot embodiment region with plausible human arms and hands.

Given an original robot frame $y_i^e$, the open-domain editor produces a human-like image $\tilde{x}_i^h$. We manually inspect the edited candidates and keep only examples where the non-embodiment content remains well aligned. In particular, we discard cases with background changes, object displacement, camera-viewpoint drift, incorrect hand-object contact, inconsistent geometry, or severe editing artifacts. After this curation step, each target embodiment yields approximately 300 high-quality pseudo pairs.

The curated dataset for embodiment $e$ is written as

$$\mathcal{D}^{(e)} = \left\{ \left( \tilde{x}_i^h, y_i^e, p_i^{(e)} \right) \right\}_{i=1}^{N_e}. \tag{1}$$

Here, $\tilde{x}_i^h$ is the human-like image synthesized from the robot frame, $y_i^e$ is the original robot image, and $p_i^{(e)}$ is the editing instruction that specifies the target robot embodiment. Although the pseudo pair is constructed through robot-to-human editing, the supervised editor is trained in the inverse direction, from $\tilde{x}_i^h$ to $y_i^e$. This converts the strong human-generation prior of a general image-editing model into aligned supervision for learning embodiment-specific human-to-robot replacement.

#### 2.1.2 Paired offline augmentation

To increase appearance diversity while preserving the correspondence between source and target images, we apply synchronous offline augmentation to each pseudo pair $(\tilde{x}_i^h, y_i^e)$. All geometric and photometric transformations are applied identically to both images, so that the embodiment region, object layout, and background remain spatially aligned after augmentation. We use horizontal flipping, random crop-and-resize, and brightness perturbation. The crop parameters and brightness factors are sampled once per pair and shared by the source and target images. With flipped and non-flipped variants, three crop settings, and three brightness settings, each curated pair is expanded into

$$N_{\text{aug}} = 2 \times 3 \times 3 = 18. \tag{2}$$

We further remove fisheye-camera samples from the final split to reduce strong lens distortion and improve geometric consistency.

#### 2.1.3 Embodiment editor fine-tuning

We fine-tune Qwen-Image-Edit (Wu et al., 2025; Qwen Team, 2025) as the embodiment editor. The model takes the human-like source image $\tilde{x}_i^h$ and the target-embodiment prompt $p_i^{(e)}$ as conditions, and is supervised to reconstruct the corresponding robot image $y_i^e$. We full-tune the diffusion transformer while using the pretrained tokenizer, image processor, text encoder, and VAE from the base image-editing model. The training objective is a supervised flow-matching loss:

$$\mathcal{L}_{\text{edit}} = \mathbb{E}_{(\tilde{x}_i^h, y_i^e, p_i^{(e)}), t, \epsilon} \left[ \left\| v_\theta \left( z_t, t, p_i^{(e)}, \tilde{x}_i^h \right) - v_t^\star \right\|_2^2 \right]. \tag{3}$$

Here, $z_t$ is the noisy latent of the robot target image, and $v_t^\star$ is the flow-matching target. In practice, we train an embodiment-specific editor for each target robot. This allows the model to learn the morphology, material appearance, and spatial placement of each robot body while keeping the same pseudo-paired training formulation.

#### 2.1.4 First-frame editing for world-model generation

At inference time, the trained editor is applied in the direction required by our world-model pipeline. Given the first frame of a human egocentric demonstration, we use the embodiment-specific prompt $p^{(e)}$ to replace visible human body parts with the target robot embodiment. The edited frame preserves the task scene, object state, and camera viewpoint, while specifying the desired robot morphology for the initial visual condition. This edited first frame is then used as the target-embodiment initial state for AnyWorld rollout generation.

This reverse pseudo-pair strategy is important for our setting. It avoids the need for real paired human–robot supervision, leverages the strongest prior of open-domain image editors, and converts unpaired robot

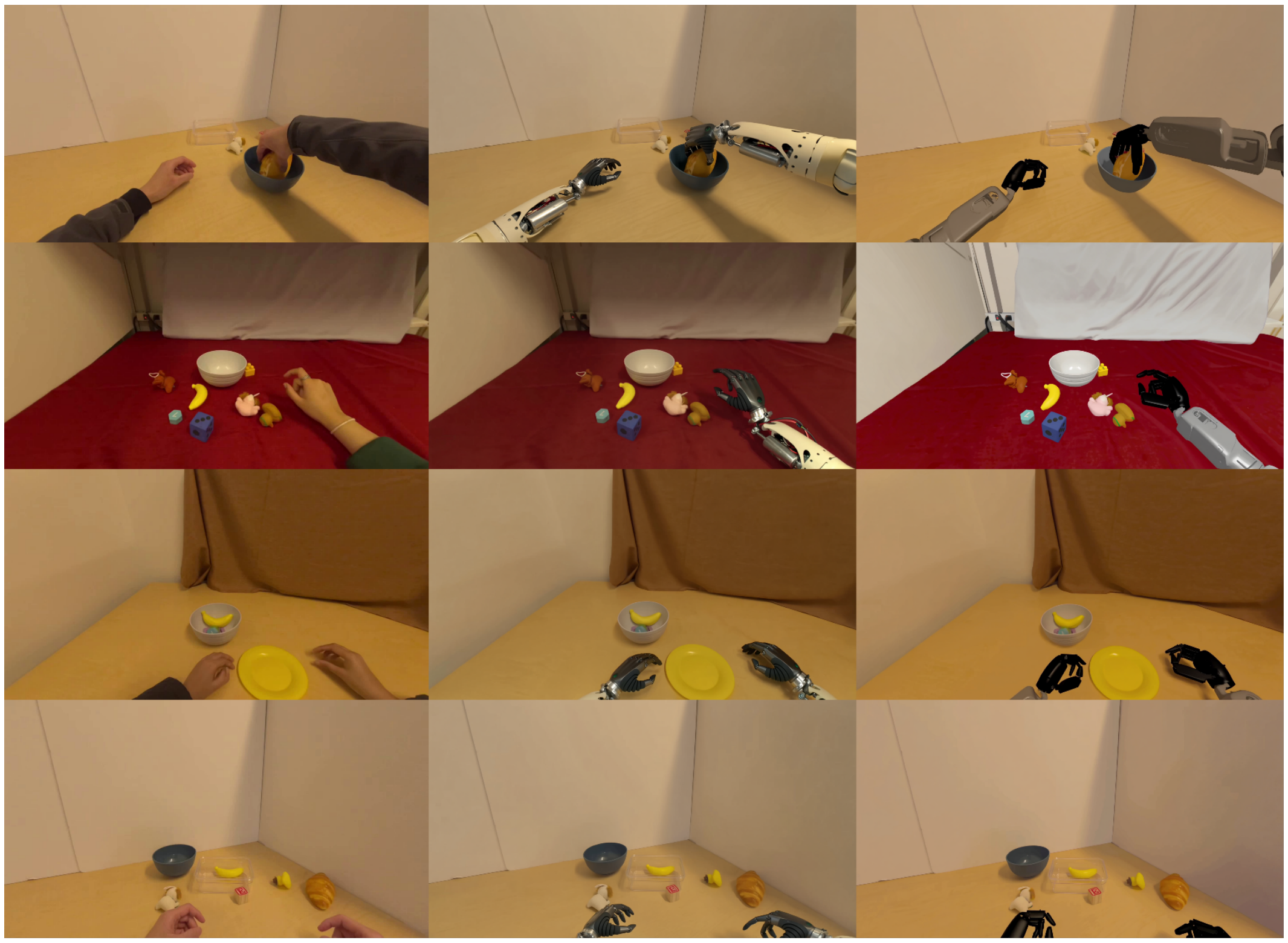

Figure 1: **Generalizable human-to-robot embodiment editing.** Each row shows one human egocentric input and its edited results for different target robot embodiments, including IRON and RoboCasa GR1. The learned editor preserves the scene layout, object state, camera viewpoint, and hand-object geometry while accurately replacing human hands with target robot morphology. Across the examples, the edited robot hands follow the original hand position and fine-grained finger articulation. The last row further shows that the editor remains effective when the source hand is only partially visible.

frames into aligned training data for human-to-robot embodiment editing. As a result, the image editor provides a practical bridge between human egocentric demonstrations and robot-native world-model rollouts.

Figure 1 shows qualitative results of the learned embodiment editor. Across four representative examples, the editor accurately replaces human hands with both IRON and RoboCasa GR1 embodiments while preserving the original hand location, interaction geometry, and surrounding scene. The edited robot hands also maintain fine-grained pose details, such as finger bending and contact-oriented hand configuration. Notably, the last example contains only a partially visible hand, yet the editor still produces a plausible and well-aligned robot embodiment, indicating robustness to partial observations and occlusions.

### 2.2 World Model Training Details

**Backbone and fine-tuning strategy.** We build AnyWorld on top of the Wan 2.1 14B video generation backbone (Wan Team, 2025; Alibaba PAI, 2025). Rather than using parameter-efficient adaptation, we perform full fine-tuning of the world-model backbone and the newly introduced conditioning modules. This choice is important for our setting because cross-embodiment rollout generation requires the model to jointly adapt its motion prior, camera-following behavior, and embodiment-conditioned appearance generation. All training stages use the same factorized interface described in the main paper: an action-control video, a camera trajectory condition, and an embodiment-aware text condition.

**Human interaction pretraining.** We first pretrain the model on large-scale human egocentric manipulation videos from EgoDex (Hoque et al., 2025). Each training sample is converted into a 115-frame clip

Table 1: **World-model training setup.** We use full fine-tuning in both stages. Human pretraining learns general egocentric interaction priors, while mixed-embodiment fine-tuning grounds the model in target robot embodiments.

| Stage | Data | Clip length | Hardware | Steps |
|---|---|---|---|---|
| Human pretraining | EgoDex | 115 frames | 32 H200 | 30K |
| Mixed-embodiment fine-tuning | EgoDex + robot data | 115 frames | 8 H200 | 5K |

with aligned action-control video, camera trajectory, first-frame condition, and structured text description. This stage is designed to learn general contact-rich human interaction priors before exposing the model to robot embodiments. We train the model for 30K steps on 32 NVIDIA H200 GPUs. The pretraining stage provides a strong action-camera conditioned world prior over egocentric manipulation, which is later reused during mixed-embodiment fine-tuning.

**Mixed-embodiment fine-tuning.** After human pretraining, we perform mixed-embodiment fine-tuning on unpaired human and robot videos. This stage is also conducted with full fine-tuning, and is trained for 5K steps on 8 NVIDIA H200 GPUs. The mixed data includes EgoDex human clips and robot-domain clips from target embodiments such as IRON and RoboCasa GR1 (Nasiriany et al., 2024). To preserve the generalizable human-world knowledge learned during pretraining while grounding the model in robot appearances, we maintain a 1:1 ratio between human and robot data within each epoch. Importantly, the human subset is re-sampled across epochs instead of being fixed. This allows the model to retain exposure to as much diverse human interaction data as possible while repeatedly observing the robot-domain data needed for embodiment grounding. In practice, this sampling strategy stabilizes robot embodiment generation without catastrophically overwriting the human interaction prior acquired during pretraining.

**Structured embodiment tags.** For each training video, we construct a structured caption that explicitly separates the scene, action, and acting subject. This structured description reduces ambiguity in the text condition and allows the same action and scene description to be reused across different embodiments by replacing only the subject phrase. For example, an EgoDex clip uses a subject phrase such as "a human", an IRON clip uses "an IRON humanoid robot", and a RoboCasa GR1 clip uses "a RoboCasa GR1 robot".

The resulting caption has the following form:

$$\tau = \mathrm{concat}\left(\tau_{\mathrm{scene}}, \tau_{\mathrm{action}}, \tau_{\mathrm{subject}}\right), \tag{4}$$

where $\tau_{\mathrm{scene}}$ describes the environment and object layout, $\tau_{\mathrm{action}}$ describes the manipulation behavior, and $\tau_{\mathrm{subject}}$ specifies the acting embodiment.

**Embodiment binding through cross-attention.** The structured caption is encoded by the text encoder and injected into the video diffusion transformer through cross-attention. We find this simple embodiment-tagging mechanism to be sufficient for binding identity information when combined with the target first-frame condition. Therefore, we do not introduce an additional explicit embodiment classifier, identity loss, or separate embodiment-token module. This design keeps the conditioning interface simple: the first frame specifies the visual state and local geometry, while the structured embodiment tag specifies the semantic identity of the acting body.

### 2.3 Policy Adaptation Details

**VLA policy baseline.** We use UniT (Chen et al., 2026) as our VLA policy baseline. UniT is pretrained on the EgoDex (Hoque et al., 2025) `basic_pick_place` subset, which contains 27,419 trajectories, for 80K steps. This provides a strong human-video-pretrained policy initialization before target-robot adaptation. Our goal is not to modify the UniT architecture, but to test whether human experience remains useful after being re-embodied into the target robot domain.

**Intermediate adaptation with re-embodied human videos.** During adaptation to each target robot domain, including IRON and RoboCasa GR1, we augment the robot-domain training data with our generated re-embodied EgoDex rollouts. These rollouts are produced by first converting the human first frame

into the target robot embodiment and then using AnyWorld to generate the corresponding robot-domain video rollout. The generated data preserves the original human interaction structure, camera motion, and task semantics, while presenting the observation in the target robot embodiment. We use this mixed data for an intermediate adaptation stage of 1K training steps. The training recipe, action format, and policy objective are kept the same as in UniT, so the generated data is introduced only as additional target-domain visual experience.

**Final robot-only fine-tuning.** After the intermediate adaptation stage, we further fine-tune the policy using only real target-robot data for 500 steps. This final robot-only stage adapts the policy to the deployment embodiment and removes possible residual distribution mismatch introduced by generated rollouts. For a fair comparison, the baseline and our method use the same UniT initialization, the same target-robot data, and the same final robot-only fine-tuning protocol. The only difference is that our method adds re-embodied human videos during the intermediate adaptation stage. Therefore, the performance gain directly reflects the benefit of converting human egocentric experience into robot-domain training experience.

### 2.4 Targeted Human-to-IRON Visual–Action Pair Construction

**Joint visual and action transfer.** The targeted closed-loop experiments require the visual observation and action label to be transferred together. Given a human demonstration, AnyWorld first converts the visible human operator into IRON and edits the relevant object configuration while preserving task semantics, camera viewpoint, background layout, and interaction timing. The corresponding human wrist trajectory is independently calibrated into the IRON action space. Each training item therefore takes the form

$$\left(\hat{V}_{1:T}^{\mathrm{IRON}},\, \ell,\, \widetilde{A}_{1:T}^{\mathrm{IRON}}\right), \tag{5}$$

so that the robot embodiment visible in the generated video is consistent with the morphology represented by the action supervision.

**Morphology-calibrated action transfer.** We resample the source human trajectory from its native 16 FPS to 30 FPS and construct a continuous 50-step action sequence. Position is interpolated in Euclidean space, while rotation is interpolated with SO(3) spherical linear interpolation. We then express the motion as donor-relative local deltas, project the active side into the corresponding IRON-hand PCA action subspace, calibrate position, velocity, and rotation scales, and clip the result to the valid IRON dynamic range. The trajectory is re-anchored to the shared initial robot state $S_0$ of the paired IRON sample and finally converted by the same RelativeActionTransform used at deployment. The target side follows the calibrated continuous trajectory, whereas the non-target side is kept close to its initial state.

**Completion-shortcut intervention.** The robot-only baseline is able to grasp and place a banana in the standard single-banana setting, but produces only small local motion when one banana is already inside the box. We therefore intervene on the missing visual state by adding a banana inside the box while retaining another valid target outside. Human interactions are re-embodied and re-scened into this configuration, producing IRON-domain observations paired with calibrated task actions. This controlled augmentation targets the false completion prior without changing the policy architecture or retraining the frozen visual-language backbone.

**Instruction-conditioned dual-target pairs.** For target selection, we construct approximately counterfactual pairs with the same symmetric dual-target visual state but opposite language instructions and corresponding actions:

$$(I_{\mathrm{dual}},\, \ell_{\mathrm{left}},\, A_{\mathrm{left}}), \qquad (I_{\mathrm{dual}},\, \ell_{\mathrm{right}},\, A_{\mathrm{right}}), \tag{6}$$

where $\ell_{\mathrm{left}}$ is “Place the left-side banana into the box” and $\ell_{\mathrm{right}}$ is “Place the right-side banana into the box.” The visual configuration is matched, while the instruction and aligned action change together. This construction prevents the policy from solving the task through a fixed side bias and instead supervises the binding among language, target location, and robot action. Based on the active side in the source videos, the final targeted set contains 26 left-target samples and 38 right-target samples.

## 2.5 Metric Details

We use lightweight automatic metrics to evaluate whether generated rollouts follow the specified action, camera, and embodiment conditions. These metrics are designed as controllability proxies rather than direct measures of physical task success. All videos are uniformly sampled to 16 frames and resized to a fixed resolution before evaluation.

**ActionAlign.** ActionAlign measures whether the local motion in the generated video follows the input action-control video. We obtain an action region from the rendered control skeleton by thresholding and dilation. Dense optical flow is computed for both the control signal and the generated video. To reduce the effect of egocentric camera motion, we estimate the global background motion from regions outside the action mask and subtract it from the generated flow. The score is the average positive cosine similarity between the residual generated flow and the control-skeleton flow inside the valid action region:

$$\text{ActionAlign} = \frac{1}{|\mathcal{T}_a|} \sum_{t \in \mathcal{T}_a} \max\left\{0, \cos\left(R_t, S_t\right)\right\}. \tag{7}$$

where $R_t$ denotes the background-compensated generated flow and $S_t$ denotes the control-skeleton flow. Higher values indicate better action-following consistency.

**CameraAlign.** CameraAlign evaluates whether the generated video follows the camera motion of the source video. For both the source and generated videos, we estimate dense optical flow and use the median background flow as a proxy for global camera motion. When an action-control mask is available, the action region is excluded to avoid contamination from hand or robot motion. The metric is the average positive cosine similarity between the source and generated camera-motion vectors:

$$\text{CameraAlign} = \frac{1}{|\mathcal{T}_c|} \sum_{t \in \mathcal{T}_c} \max\left\{0, \cos\left(c_t^{\text{gen}}, c_t^{\text{src}}\right)\right\}. \tag{8}$$

where $c_t^{\text{src}}$ and $c_t^{\text{gen}}$ are the estimated background-motion vectors of the source and generated videos. This metric mainly reflects whether the generated rollout follows the same global pan, tilt, or translation direction as the source camera trajectory.

**EmbodAcc.** EmbodAcc measures whether the generated video matches the requested embodiment. We uniformly sample up to eight frames from each generated video and crop the lower image region, where hands or robot arms are most likely to appear. We then use CLIP (Radford et al., 2021) to compare each cropped frame with human and robot prompt sets. For robot targets, a frame is counted as correct when its similarity to robot prompts is higher than its similarity to human prompts; for human targets, the decision is reversed. The metric is the average frame-level correctness:

$$\text{EmbodAcc} = \frac{1}{|\mathcal{I}|} \sum_{i \in \mathcal{I}} \mathbf{1}\left[\text{Embodiment}(\hat{Y}_i) = e_{\text{target}}\right], \tag{9}$$

where $\hat{Y}_i$ is the cropped generated frame and $e_{\text{target}}$ is the requested embodiment. EmbodAcc is therefore a binary embodiment-recognition accuracy rather than a continuous visual-fidelity score.

**Aggregation.** For each method, we report the mean score over all evaluated videos. When applicable, we also compute the standard error of the mean across videos. Invalid samples, such as frames without reliable motion estimates or unavailable CLIP scores, are excluded from the corresponding metric average.

## 2.6 Physically Consistent Outcome Simulation

We further evaluate whether AnyWorld can simulate both successful and failed interaction outcomes under different root-motion conditions. As shown in Fig. 2, we use the same initial scene and vary the root motion while keeping the scene layout, object configuration, and embodiment fixed. This setting tests whether the model can respond to changes in embodied motion, rather than merely generating visually plausible rollouts.

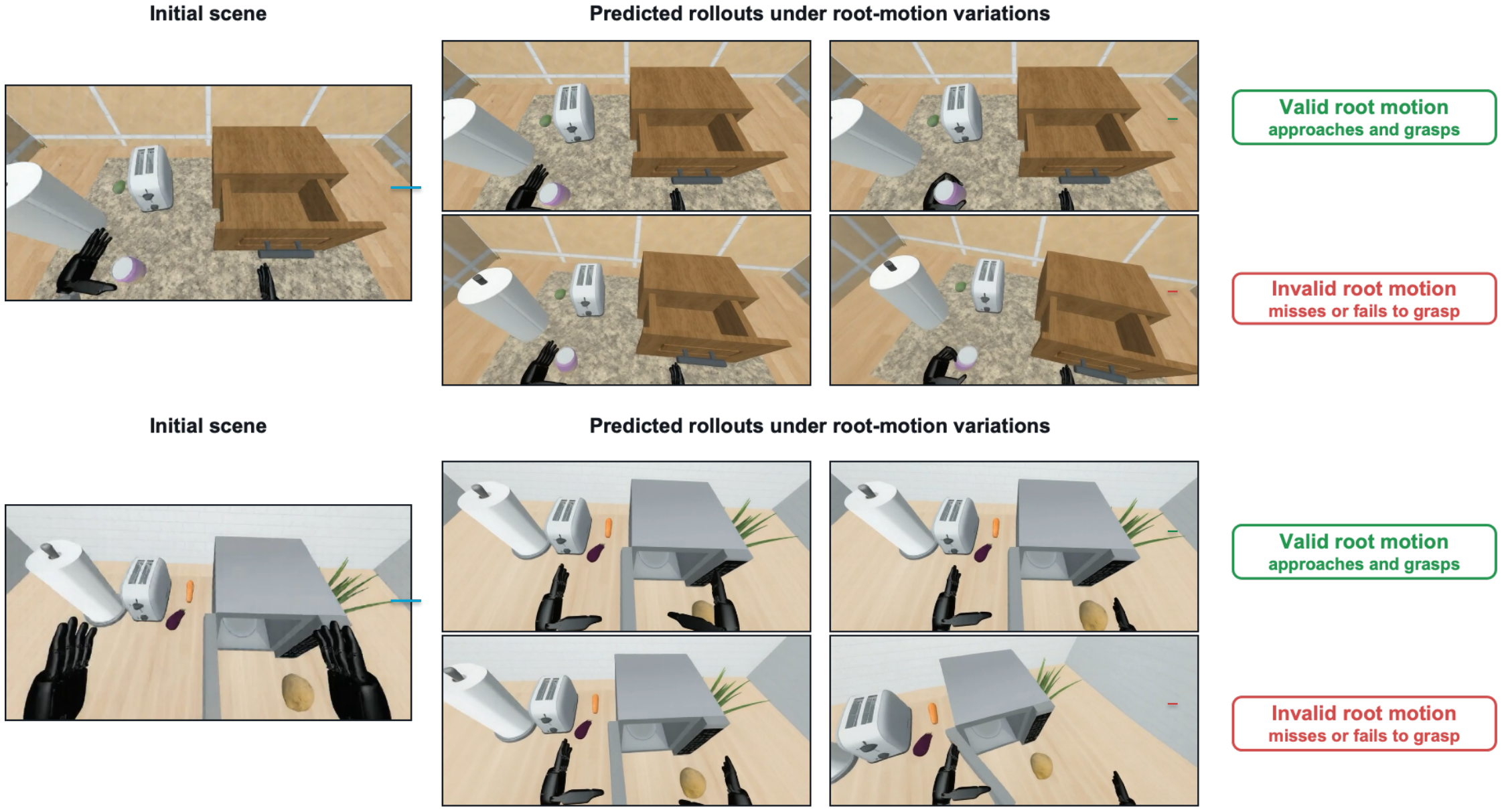


Figure 2: **Physically consistent outcome simulation under root-motion variations.** Given the same initial scene, AnyWorld predicts different interaction outcomes when the root motion changes. Valid root motions lead the robot to approach and grasp the target object, while invalid root motions produce plausible failed interactions where the robot misses or fails to grasp.

Across the two examples, AnyWorld generates outcomes that are consistent with the imposed root motion. When the root motion brings the robot close to the target object, the generated rollout preserves the approach behavior and produces a successful grasp. In contrast, when the root motion is spatially misaligned with the target, the model does not force a successful interaction; instead, it predicts a failure case in which the robot misses or cannot complete the grasp. This suggests that the model captures the dependence between global motion, object reachability, and downstream contact outcome.

This property is important for using generated rollouts as robot experience. A useful world model should not only synthesize realistic videos, but also reflect how changes in motion conditions affect interaction success. The results indicate that AnyWorld can model both positive and negative outcomes under controlled root-motion variations, providing more informative rollouts for downstream policy learning and evaluation.



\end{document}